\documentclass[10pt,twocolumn,letterpaper]{article}

\usepackage{wacv}
\usepackage{times}
\usepackage{epsfig}
\usepackage{graphicx}
\usepackage{amsmath}
\usepackage{amssymb}
\usepackage{multirow}
\usepackage{array,balance}

\wacvfinalcopy 

\def\wacvPaperID{1102} 

\ifwacvfinal\fi
\begin{document}

\title{Plug-and-Play Rescaling Based Crowd Counting in Static Images}

\author{Usman Sajid,\hspace{2cm} Guanghui Wang\\
Electrical Engineering and Computer Science, The University of Kansas, Lawrence, KS, USA 66045\\
{\tt\small \{usajid, ghwang\}@ku.edu}
}

\maketitle
\ifwacvfinal\thispagestyle{empty}\fi

\begin{abstract}
Crowd counting is a challenging problem especially in the presence of huge crowd diversity across images and complex cluttered crowd-like background regions, where most previous approaches do not generalize well and consequently produce either huge crowd underestimation or overestimation. To address these challenges, we propose a new image patch rescaling module (PRM) and three independent PRM employed crowd counting methods. The proposed frameworks use the PRM module to rescale the image regions (patches) that require special treatment, whereas the classification process helps in recognizing and discarding any cluttered crowd-like background regions which may result in overestimation. Experiments on three standard benchmarks and cross-dataset evaluation show that our approach outperforms the state-of-the-art models in the RMSE evaluation metric with an improvement up to $10.4\%$, and possesses superior generalization ability to new datasets.
\end{abstract}

\section{Introduction}
\label{intro123}
Deep learning has achieved significant progress in many computer vision applications, like image classification \cite{gao2016novel, cen2019boosting}, object detection \cite{ma2019mdfn, li2019object}, face recognition \cite{cen2019dictionary}, depth estimation \cite{he2018learning, he2018spindle}, image translation \cite{xu2019toward, xu2019adversarially}, and crowd counting \cite{sajid2019zoomcount}. Crowd counting plays a vital role in crowd analysis applications such as better management of political rallies or sports events, traffic control, safety and security, and avoiding any political point-scoring on crowd numbers \cite{zhang2017fcn}. In addition to crowd estimation, the same methods can also be applied to other fields like the counting of animals, crops, and microscopic organisms \cite{arteta2016counting,sam2017switching}.

Automated crowd counting comes up with different challenges including large perspective, huge crowd diversity across different images, severe occlusion and dense crowd-like complex background patterns. Recent methods mostly employ deep convolutional neural networks (CNNs) to automate the crowd counting process. These approaches can be categorized as Detection based, Direct regression based, and Density map estimation based methods. Detection based methods use CNN based object detectors (e.g. Faster-RCNN \cite{girshick2015fast}, YOLO \cite{redmon2016you}) to detect each person in the image. The final crowd count is the sum of all detections. This idea does not generalize well for high-density crowd images, where detection fails due to very few pixels per head or person. Direct regression based methods \cite{huang2017densely, sajid2019zoomcount} learn to directly regress crowd count from the input image. These methods alone cannot handle huge crowd diversity and thus lack generalization. Density map estimation based methods \cite{sam2017switching,cascadedmtl,zhang2018crowd,liu2019context,wan2019residual,shi2019counting,xiong2019open,jiang2019crowd} estimate crowd density value per pixel instead of the whole input image. Most current state-of-the-art methods belong to this category due to their better and effective performance, however, limitations related to density map estimation per pixel pose a huge challenge \cite{ranjan2018iterative} due to large variations in crowd number across different images.

\begin{figure}[t]
\label{fig:figSecond}
	\begin{minipage}[b]{0.45\columnwidth}
		\begin{center}
			\centerline{\includegraphics[width=0.695\columnwidth]{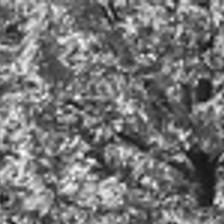}}
			\centerline{\footnotesize{Ours=0, DR=61, DME=52}}
		\end{center}
	\end{minipage}
		\begin{minipage}[b]{0.08\columnwidth}
		\begin{center}
			\centerline{\footnotesize{ }}
		\end{center}
	\end{minipage}
	\begin{minipage}[b]{0.45\columnwidth}
		\begin{center}
			\centerline{\includegraphics[width=0.695\columnwidth]{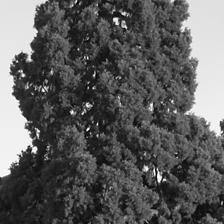}}
			\centerline{\footnotesize{Ours=0, DR=27, DME=53}}
		\end{center}
	\end{minipage}
			
    \vspace{-3mm}
	\caption{\footnotesize{Current Direct Regression (DR) and Density-map estimation (DME) \cite{idrees2018composition} based methods overestimate in case of even very small ($224 \times 224$) size cluttered crowd-like background regions in images, as they face difficulty in recognizing and discarding such complex patterns.
	}}
    \vspace{-4mm}
    
\end{figure}

Multi-column CNN (MCNN) model \cite{zhang2016single} is a three-column density-map estimation based network, that uses different filter sizes in each of its columns to account for multiple scales. Each branch is specialized in handling the respective scale. These columns are eventually concatenated at the end to output the final crowd estimate. Similarly, another state-of-the-art model, named Switch-CNN \cite{sam2017switching}, deploys a hard switch to select one of three specialized crowd count regressors accordingly for the input image. Each count regressor is specialized to handle and focus on respective crowd density. Non-switch and single-column based models \cite{zhang2018crowd} are also being designed to solve the counting issues, but they lack the ability to generalize well on huge crowd diversity across different images, and thus, result in either high over-estimation or under-estimation.

Another key issue with these methods is their noticeable inability and lack of focus towards detecting and discarding any cluttered crowd-like background region or patch in the image that may cause huge crowd over-estimation. As shown in Fig. 1, current methods do not detect these $224 \times 224$ cluttered crowd-like background regions in the images and thus result in crowd over-estimation. This problem would scale-up quickly with more such regions occurring regularly in the images.

To address the aforementioned major issues, we propose a simple yet effective image patch rescaling module (PRM) and three new crowd counting frameworks employing the plug-and-play PRM module. These frameworks range from a modular approach to multi-task end-to-end networks. The lightweight PRM module addresses the huge crowd diversity issue efficiently and effectively, and also appears as a better alternative to the recent computationally heavy multi-column or multiple specialized count regressors based architectures. In the proposed frameworks, high-frequency crowd-like background regions also get discarded that may cause huge crowd overestimation otherwise. The main contributions of this work are as follows:

\begin{itemize}\setlength\itemsep{-0.1em}

  \item We propose a conceptually simple yet effective and plug-and-play based patch rescaling module (PRM) to address the major huge crowd diversity issue in crowd counting problems.
  \item We also propose three new and independent crowd counting frameworks that utilize the lightweight PRM module instead of computationally expensive recent multi-column or multi-regressor based architectures.
  \item Extensive experiments on three benchmark datasets show that our approach outperforms the state-of-the-art methods in terms of RMSE evaluation metric with the improvement up to $10.4\%$. Cross-dataset evaluation also demonstrates the better generalization ability of the proposed PRM module and crowd counting schemes relative to similar state-of-the-art methods.

\end{itemize}

\section{Related Work}
Different crowd counting methods have been proposed over the time to address the key problems like huge crowd diversity, severe occlusions, cluttered crowd-like background regions, and large perspective changes. Classical approaches can be categorized into two classes: Counting by detection and counting by regression. Count by detection classical methods first detect individual persons in the image \cite{wu2005detection,wang2011automatic,ge2009marked,li2008estimating} using handcrafted features. Then, the final image count is obtained by the summation of all detections. These detectors fail quite drastically in the case of high-density crowd because of few pixels per person. Classical regression based methods \cite{chan2009bayesian,chen2012feature,ryan2009crowd} learn to directly map the image local patch to the crowd count. They yield better performance, however, they still suffer from lack of generalization on reasonable crowd diversity range.

\begin{figure*}
	\begin{minipage}[b][][b]{1.05\columnwidth}
		\begin{center}
			\centerline{\includegraphics[width=1\columnwidth]{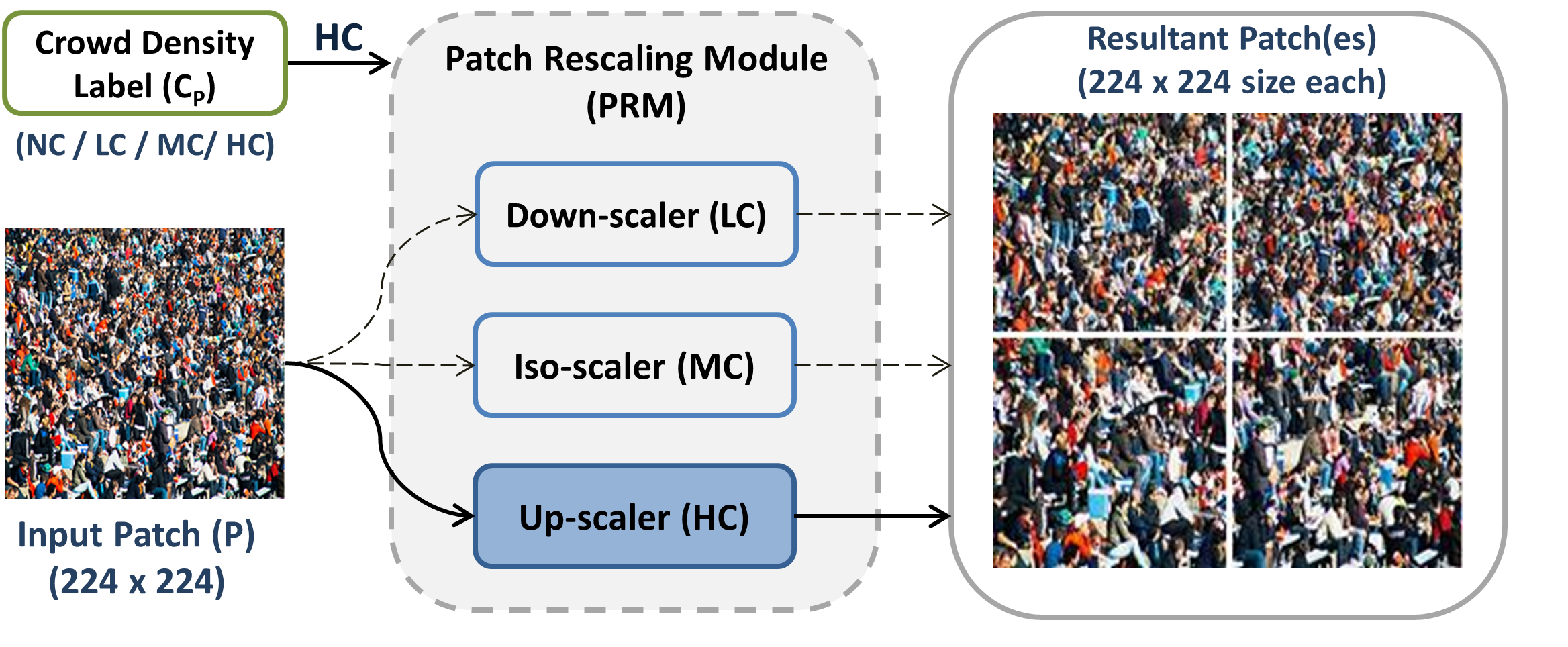}}
			\centerline{\footnotesize{(a) PRM Module}}
		\end{center}
	\end{minipage}
		\begin{minipage}[b][][b]{0.44\columnwidth}
		\begin{center}
			\centerline{\includegraphics[width=1\columnwidth]{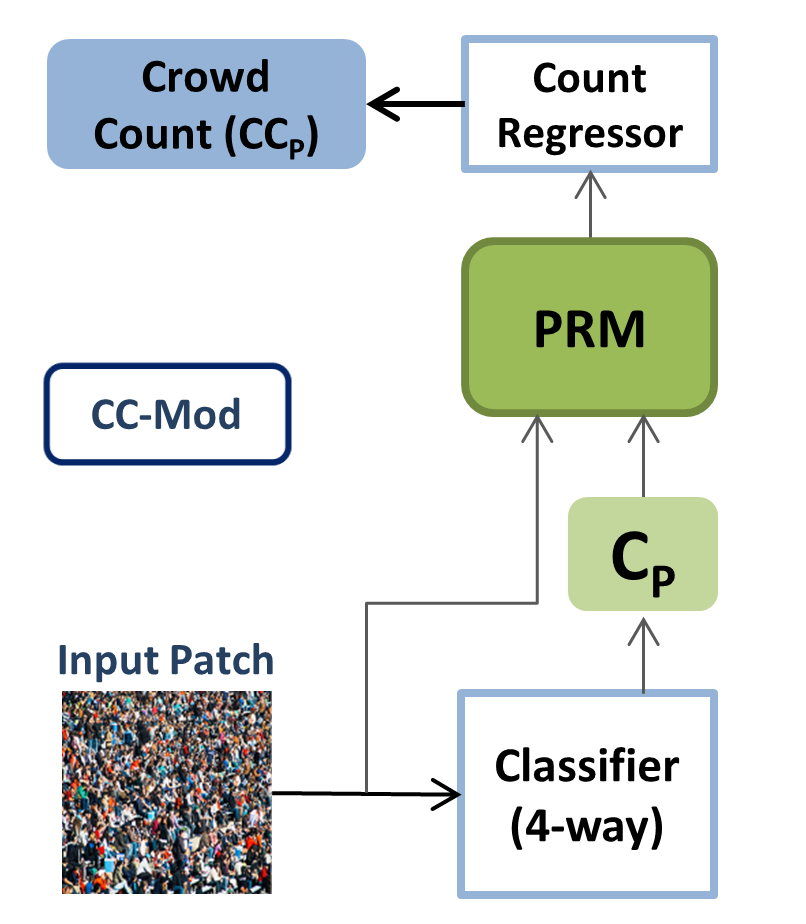}}
			\centerline{\footnotesize{(b) Modular Scheme (CC-Mod)}}
		\end{center}
	\end{minipage}
	\begin{minipage}[b][][b]{0.60\columnwidth}
		\begin{center}
			\centerline{\includegraphics[width=1\columnwidth]{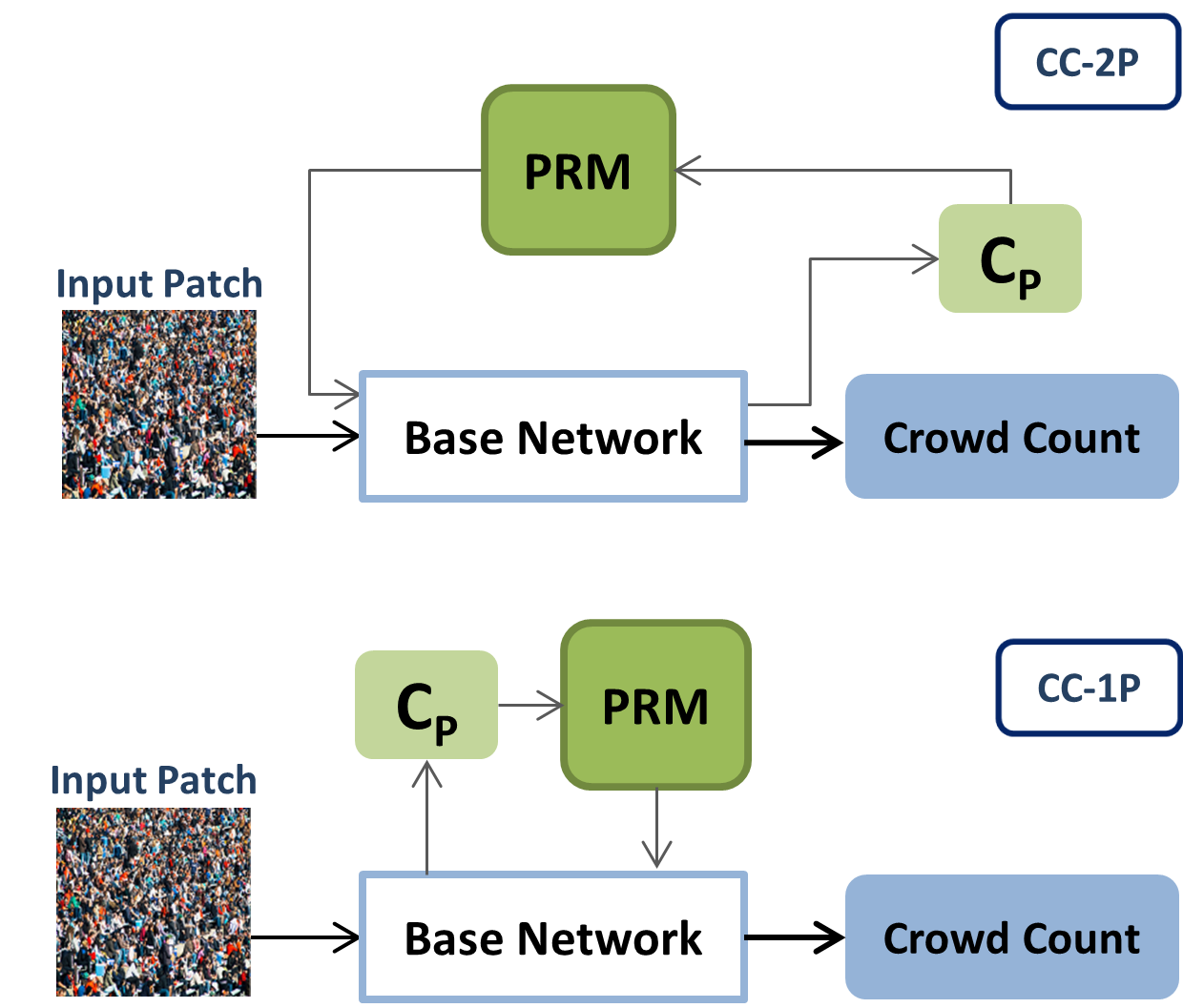}}
			\centerline{\footnotesize{(c) End-to-End Networks (CC-1P, CC-2P)}}
		\end{center}
	\end{minipage}
				
    \vspace{-2mm}
	\caption{\footnotesize{(a) \textbf{PRM Module.} Based on the prior estimated crowd-density class ($C_P$), the PRM module rescales the input patch $P$ (when $C_P=HC\ or\ LC$) using one of its rescaling operations (\textit{Up-scaler or Down-scaler}) and generates $4$ or $1$ new rescaled patch(es) respecively. The $MC$ labeled patch bypasses any rescaling (\textit{Iso-scaler}). (b) \textbf{CC-Mod}. In the modular crowd counting scheme, the input patch is first classified 4-way (NC, LC, MC, HC), followed by passing through the PRM and then through the regressor for final patch crowd count ($CC_P$). (c) \textbf{CC-1P, CC-2P.} These crowd counting networks couple the PRM module with the base network to address the huge crowd diversity issue amid giving better performance (Architectures detailed in Figs. \ref{fig:cc_2p} and \ref{fig:cc_1p}). }} 
	\label{fig:proposed_schemes}
    \vspace{-4mm}
\end{figure*}

CNN based methods have been widely used nowadays for crowd counting. They are of three types: Counting by detection, by regression, and density-map estimation based methods. In detection based methods, CNN detectors (like YOLO, Faster-RCNN \cite{girshick2015fast,redmon2016you}) detect each person in the image, followed by the sum of all detections to yield final crowd estimate. These methods fail in the case of high occlusion and crowd. Regression based methods \cite{walach2016learning,wang2015deep} learn to directly regress the crowd count. These regressors alone do not generalize well for the huge crowd diversity range. 

The density-map estimation methods \cite{zhang2016single,sam2017switching,cascadedmtl,zhang2018crowd,liu2019context,wan2019residual,shi2019counting,xiong2019open,jiang2019crowd} estimate crowd density value per pixel. The final count is obtained from the sum of all pixel density values. These methods are widely used recently with better performance and focus on multi-scale or diverse crowd-density range handling. Zhang \textit{et al.} \cite{zhang2016single} proposed a three-column architecture, where each column caters respective scale using different filter sizes, followed by a fusion of the columns to produce final crowd density-map. Similar to this idea, Switch-CNN \cite{sam2017switching} utilized three specialized count regressors to cater three different scales. Each input image routes to one of these regressors using a switch CNN-based classifier. Cascaded-mtl \cite{cascadedmtl} generated density-map by first classifying image 10-way prior to actual density-map production. Recently, \textit{Zhang et al.} \cite{zhang2018crowd} proposed a scale-adaptive network, which employs a single backbone structure and the same filter size and combines different layers feature maps to handle the multi-scale variation. Ranjan \textit{et al.} \cite{ranjan2018iterative} proposed a two-column network where first low-resolution based column feature maps are combined with high-resolution input column to generate final density map. These methods still struggle to handle the huge crowd diversity and thus do not generalize well. They also overestimate for the cluttered background regions in the images.

In order to handle multi-scale or huge crowd diversity variation and cluttered background regions, we propose a conceptually straightforward yet effective solution, which explores basic rescaling operations and three different schemes utilizing these operations to address the above issues and generalize well on even cross-dataset setting.

\section{Proposed Method}
Our method focuses on addressing huge crowd diversity within as well as across different images and the presence of cluttered crowd-like background regions in these images. As shown in Fig. \ref{fig:proposed_schemes}(a), the proposed Patch Rescaling Module (PRM) comprises of three straightforward operations i.e. \textit{Up-scaler, Down-scaler and Iso-scaler}. The input image patch uses one of these operations to adjust its scaling accordingly, depending on its crowd density level. This lightweight rescaling process helps in addressing the crowd diversity issue efficiently and effectively. Next, we propose three new and different crowd counting schemes that employ the plug-and-play PRM module as shown in Fig. \ref{fig:proposed_schemes}. These frameworks include a modular approach (Fig. \ref{fig:proposed_schemes}b) and two end-to-end (Fig. \ref{fig:proposed_schemes}c) networks. The modular framework uses the PRM block in between the independent classification and regression modules, while end-to-end multi-task networks utilize the PRM to facilitate the base network for better and efficient performance. Both PRM and the proposed frameworks are detailed in the following text.

\subsection{Patch Rescaling Module (PRM)}
The PRM module, as shown in Fig. \ref{fig:proposed_schemes}a, is used to rescale the input patch by utilizing one of the two rescaling operations, namely \textit{Up-scaler} and \textit{Down-scaler}. PRM module selects the appropriate rescaling operation based on the crowd density level ($C_P$), which is computed prior to the PRM  module usage by the 4-way classification (\textit{no-crowd (NC), low-crowd (LC), medium-crowd (MC), high-crowd (HC)}). Crowd patches, classified as \textit{LC or HC}, pass through the \textit{Down-scaler} or \textit{Up-scaler} operation, respectively. The $MC$ labeled input patch bypasses the PRM without any rescaling (denoted by \textit{Iso-scaler}). \textit{NC} labeled input patch is automatically discarded without any PRM processing as it is a background region with zero people count. Every patch, coming out of the PRM module, will have the same fixed $224 \times 224$ size. By using the right scale for each input patch, the straightforward PRM module addresses the huge crowd diversity challenge and has been used as a plug-and-play block in different new crowd counting schemes given in Sec. \ref{cc_schemes}. Each rescaling operation is detailed below.

\textbf{Upscaling Operation (Up-scaler).}
The upscaling option is applied to the patches with high-crowd (HC) crowd. Up-scaler divides the input patch into four $112 \times 112$ patches, followed by upscaling of each new patch to $224 \times 224$ size. Intuitively, this simple operation facilitates the counting process by further dividing and zooming-in into each sub-divided part of the highly dense crowd patches separately. Consequently, it avoids the overestimation that occurs in complex multi-column architectures and multiple specialized count regressors based methods. Thus, this operation outputs four rescaled patches from the input patch.

\textbf{Downscaling Operation (Down-scaler).}
The patches that are classified as low-crowd (LC) label are subjected to downscaling operation, where the patches are first down-scaled by $2\times$  and then zero-padded to $224 \times 224$ before proceeding for further processing. Primarily, this operation helps in avoiding underestimation by using smaller area for the input patch and achieves better results without the need for any specialized or complex additional component.

\textbf{Iso-scaling block.} The image patches that are labeled as medium-density (MC) class do not require any special attention as given to LC or HC based patches, because the deep CNN based crowd counting models can handle these cases effectively without using any upscaling or downscaling operation. Thus, they are directly forwarded to the next stage for crowd estimation.

\subsection{PRM based Crowd Counting Frameworks}
\label{cc_schemes}
In this section, we discuss three independent proposed crowd counting schemes, ranging from a modular framework to two end-to-end multi-task networks. These methods address the huge crowd diversity using the PRM module as well as discard any cluttered background regions in the images. In each scheme, the input image is divided into $224 \times 224$ non-overlapping patches. Each patch then passes through that specific scheme for patch crowd count estimate. The final crowd count of the image is obtained by summing all its patches count. Each crowd counting scheme is discussed in the following subsections.

\subsubsection{Modular Crowd Counting Scheme (CC-Mod)}
The modular crowd counting framework (CC-Mod), as shown in Fig. \ref{fig:proposed_schemes}b, consists of three main components, namely Deep CNN based 4-way classifier, PRM module, and crowd count regressor. Input image gets divided into $224$ x $224$ size non-overlapping patches. Each patch is then fed to a 4-way classifier that categorizes the input patch to its appropriate crowd-density label (\textit{NC, LC, MC, HC}). Based on the assigned class label, each patch is rescaled accordingly using the PRM module before proceeding to the count regressor for the patch-wise crowd estimate. Image crowd count is finally obtained by summing all its crowd patches count. Each component has been detailed as follows.

\textbf{Crowd Density Classifier.} The goal of this module is to classify the input ($224 \times 224$) image patch into one of the four crowd density labels, namely no-crowd (NC), low-crowd (LC), medium-crowd (MC), and high-crowd (HC) crowd. The definitions of these labels are given in the next paragraph. Based on the assigned class label, each patch will then be routed to the PRB module for the rescaling operation. The $NC$ labeled patches are completely discarded without any further processing. Thus, using this specialized deep CNN classifier, we identify and discard the cluttered crowd-like background patches, which may result in huge accumulated crowd overestimation otherwise.

\textbf{Crowd-density class labels definitions.} Crowd density classifier requires the definitions of four class labels (NC, LC, MC, HC) to train and learn the 4-way classification. Since each labeled benchmark dataset contains the people localization information in the form of $(x,y)$ as the center of each person's head, we utilize this information to define the class labels and generate training patches for each class label. The ground truth crowd-density label ($C_{P(gt)}$) for the $224 \times 224$ training image patch $P$ is assigned as follows:

\begin{equation}
\label{eq1}
C_{P(gt)} =
    \begin{cases}
      NC & \text{$c_{gt}$ = 0}\\
      LC & \text{0 \textless\ $c_{gt}$ $\leq$\ 0.05 * $c_{max}$}\\
      MC & \text{0.05 * $c_{max}$ \textless\ $c_{gt}$ $\leq$\ 0.2 * $c_{max}$}\\
      HC & \text{0.2 * $c_{max}$ \textless\ $c_{gt}$}\\
    \end{cases} 
\end{equation}

\begin{figure*}

	\begin{minipage}[b]{1.0\textwidth}
		\begin{center}
			\centerline{\includegraphics[width=1.0\textwidth]{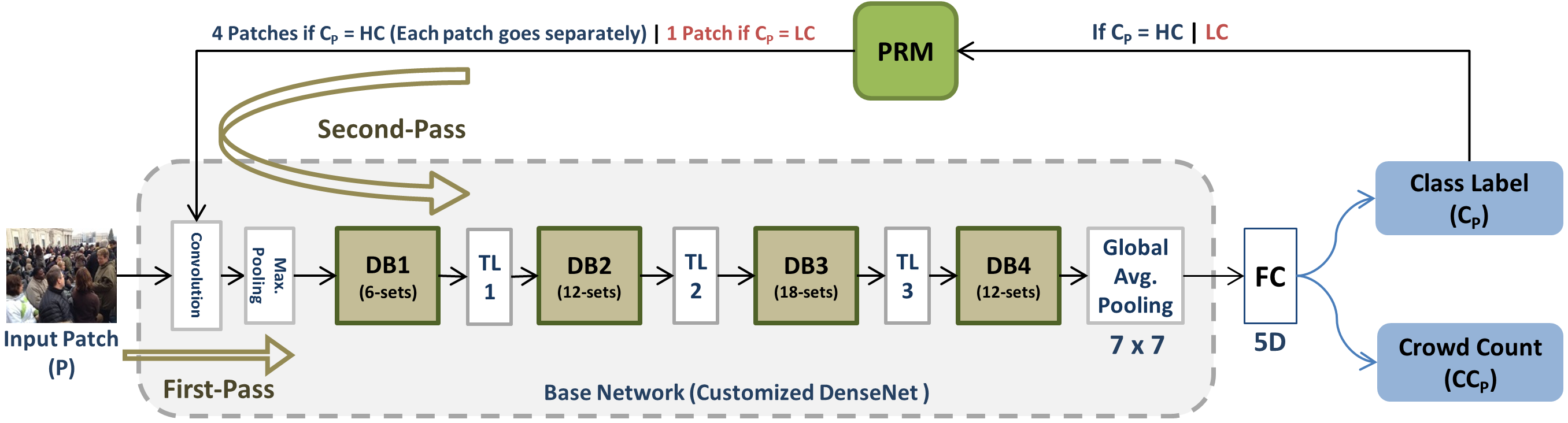}}
		\end{center}
	\end{minipage}		
    \vspace{-8mm}
	\caption{The proposed CC-2P architecture. The input patch $P$, classified as either $HC$ or $LC$ during the first-pass, passes through the base network again (second-pass) after the required PRM rescaling operation. Final patch  count ($CC_P$) is the average of both passes crowd estimates. $MC$ labeled input patch skips the second-pass without any PRM usage, and outputs the final first-pass Crowd Count ($CC_P$).}
    \vspace{-1mm}
    \label{fig:cc_2p}
    
\end{figure*}

\begin{figure*}

	\begin{minipage}[b]{1.0\textwidth}
		\begin{center}
			\centerline{\includegraphics[width=1.0\textwidth]{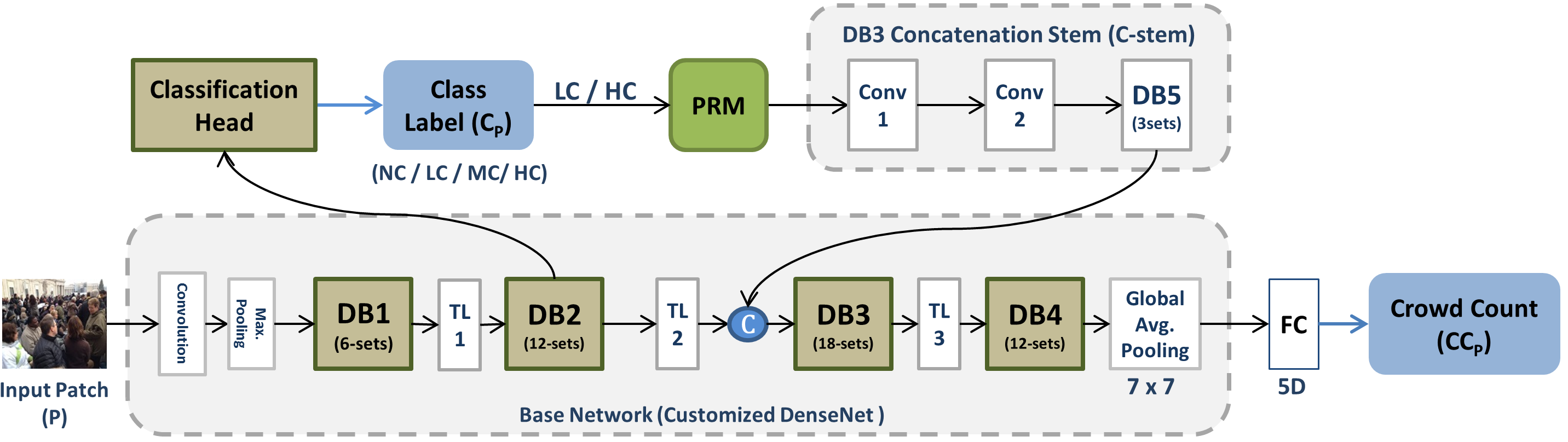}}
		\end{center}
	\end{minipage}		
    \vspace{-8mm}
	\caption{The proposed CC-1P architecture branches-out the dense block ($DB2$) output for the 4-way classification ($C_P$). The input patch $P$ then passes through the PRM for any required rescaling. Resultant patch(es) then go though the \textit{C-stem} block, followed by the channel- wise concatenation with the transition layer ($TL2$) output channels. Remaining processing finally yields the patch crowd count ($CC_P$). }
    \vspace{-3mm}
    \label{fig:cc_1p}
\end{figure*}
\noindent where $c_{gt}$ denotes the ground truth people count for the image patch $X$, $c_{max}$ stands for the possible maximum ground truth people count in any $224 \times 224$ image patch of this benchmark dataset. Image patch, containing at most $5\%$ of the maximum ground truth people count (and non-zero) is assigned with low-crowd (LC) crowd label. Similarly, a patch with actual count between $5$ to $20\%$ (including $20\%$) is classified with MC label, whereas patches containing more than $20\%$ of the maximum possible crowd count or no crowd at all are given HC or NC labels respectively. In this way, a total of $90,000$ patches ($22,500$ per class) are being generated for the classifier training during each dataset experiment separately. In all proposed schemes, we use the same class definitions for the 4-way classification.

\textbf{Classifier and Crowd Count Regressor details.} We use customized DenseNet-121 \cite{huang2017densely} based architecture as the 4-way crowd-density classifier. Empirically, we found that using only the first three dense blocks give almost the same and consistent performance for this 4-way classification task, instead of using default four dense blocks in DenseNet-121. Consequently, this reduces the learnable parameters by a huge margin ($6.95M$ to $4.80M$). At the end of the third dense block, the classification layer is composed of $7 \times 7$ global average pooling followed by the $4$D fully connected layer with a softmax 4-way classification (NC, LC, MC, HC) and cross-entropy loss.

The regressor is similar to the original DenseNet-121 architecture except that it has \{$6,12,18,12$\} sets in four dense blocks respectively instead of \{$6,12,24,16$\}. This customization significantly reduces the model parameters ($6.95M$ to $5.05M$), while performing reasonably well. In addition, the classification layer has been replaced with a single neuron to directly regress the crowd count. The mean Squared Error (MSE) has been used as the loss function for the count regressor $cr$, defined as:

\begin{equation}
\label{eq2}
Loss_{cr} = \frac{1}{n} \sum_{k=1}^{n} (F(x_k,\Theta)-y_k)^{2}
\end{equation}
where $n$ denotes the training patches per batch, $y_k$ is the actual crowd count for the input image patch $x_k$, and $F$ is the mapping function that learns the input patch $x_k$ mapping to the crowd count with weights parameters $\Theta$. 

\subsubsection{Two-Pass Crowd Counting Network (CC-2P)}
CC-2P, as shown in Fig. \ref{fig:cc_2p}, is a multi-task 4-way classification and regression based network, that employs the PRM module. The input patch goes through the base network, consisting of four dense blocks (\textit{DB1, DB2, DB3, DB4}), in the first pass to yield the crowd density class label as well as crowd number estimate. The patches, labeled as LC or HC label, proceed to the PRM module for required rescaling operation. The resultant new patch(es) then go through the base network for crowd count estimate in the second pass. Let $P$ be the original input patch and first-pass class label ($C_P$) as $LC$ or $HC$, then the final crowd count ($CC_P$) estimate for $P$ is the average of the first-pass ($fp$) and the second-pass ($sp$) crowd estimates as follows.

\setlength{\tabcolsep}{2.0pt}
\begin{table}[t]\small
	\begin{center}
	\begin{tabular}{|c|c|c|}
    \hline
 Layer & Output Size & Filters (F)\\ \hline
\multicolumn{3}{|c|}{Classification Head} \\ \hline
 & $512 \times 28 \times 28$ &     \\ \hline
 & $64 \times 28 \times 28$ &  ($1 \times 1$) conv, $64$F    \\ \hline
 & $64 \times 14 \times 14$ & ($2 \times 2$) Avg Pooling, stride $2$    \\ \hline
 & $32 \times 7 \times 7$ &  ($3 \times 3$) conv, stride 2, padding 1, $32$F   \\ \hline
 & 4D FC, softmax & -    \\ \hline
\multicolumn{3}{|c|}{DB3 Concatenation stem block (C-stem)} \\ \hline
 & $1 \times 224 \times 224$ &     \\ \hline
conv1 & $64 \times 112 \times 112$ &  ($3 \times 3$) conv, stride 2, padding 1, $64$F    \\ \hline
conv2 & $32 \times 56 \times 56$ & ($3 \times 3$) conv, stride 2, padding 1, $64$F    \\ \hline
 & $32 \times 28 \times 28$ & ($2 \times 2$) Avg Pooling, stride $2$    \\ \hline
DB3  & $128 \times 28 \times 28$ &  
$
 \begin{bmatrix}

1 \times 1\ conv \\
3 \times 3\ conv \\
        \end{bmatrix} 
$        
         $\times 3$
             \\ \hline
   & $128 \times 14 \times 14$ & ($2 \times 2$) Avg Pooling, stride $2$    \\ \hline
	\end{tabular}
	\end{center}
	    \vspace{-1mm}
	\caption{\footnotesize Configurations of the CC-1P Classification Head and the C-stem block. Each conv represents the BN-ReLU-Convolution sequence \cite{huang2017densely}.}
	\label{table:config_cc1p}
    \vspace{-6mm}
\end{table}

\begin{equation}
\label{eq3}
CC_{p} =
    \begin{cases}
      \frac{cc_{fp}\ +\ cc_{sp}}{2} & C_{p} = LC\\
      \frac{cc_{fp}+(cc_{u1}+cc_{u2}+cc_{u3}+cc_{u4})_{sp}}{2} & C_{p} = HC\\
    \end{cases} 
\end{equation}

Since the PRM produces four new upscaled patches ($u1,u2,u3,u4$) for the input patch $P$ when $C_{p} = HC$, therefore second-pass crowd count is the sum of these patches for this case. These four patches go through the network one by one during the second-pass to estimate their corresponding crowd counts. Input patch $P$ that is labeled as $MC$ in the first-pass, skips the second-pass as the PRM module has no effect on such patches (no rescaling). Also, $NC$ labeled input patch is discarded without any further processing irrespective of their crowd estimate.

\textbf{Network Details.} We use customized DenseNet-121 \cite{huang2017densely} as our base network. Empirically, it has been observed that fewer sets of ($1 \times 1$) and ($3 \times 3$) layers in the Densenet-121 deeper dense blocks ($DB3$ and $DB4$) give almost the same and consistent performance for this problem amid reducing model parameters by a significant margin ($6.95M$ to $5.05M$). Consequently, we use \{6,12,18,12\} sets instead of \{6,12,24,16\} in the four dense blocks respectively, which reduces the 121 layers deep Densenet to 101 layers. Transition layers ($TL1,TL2,TL3$) connect the dense blocks and adjust the feature maps size for the next dense block accordingly, as given in \cite{huang2017densely}. At the end of the base network, the final fully connected (FC) layer outputs the softmax based 4-way classification and regression based crowd count value. Multi-task loss ($Loss_{total}$) of CC-2P is defined as follows.

\begin{equation}
\label{eq4}
Loss_{total} = L_{reg}\ +\ L_{class} 
\end{equation}
where $L_{reg}$ is the MSE loss as defined in Eq. \ref{eq2}, $L_{class}$ is the cross-entropy loss for the softmax based 4-way labeling.

\subsubsection{Single-Pass Crowd Counting Network (CC-1P)}
The multi-task single-pass network, as shown in Fig. \ref{fig:cc_1p}, branches out the \textit{dense block 2} (DB2) output for the 4-way classification (\textit{NC, LC, MC, HC}) of the input patch. Based on the assigned class label, the input patch $P$ passes through the PRM module for any required rescaling. Patch(es), coming out of the PRM, proceed to the DB3 concatenation stem (C-stem) for the extraction of their initial feature maps that are eventually concatenated with the \textit{second transition layer} (TL2) output feature maps to serve as the input to the DB3 Finally, the global average pooling is being done on DB4 output channels followed by a single neuron to directly regress the input patch crowd count. The configurations of classification head and C-stem are shown in Table \ref{table:config_cc1p}. Base network is the same as used in CC-2P except that the \textit{compression factor} ($\theta$) for \textit{second transition layer} (TL2) has been set to $0.25$ (instead of standard DenseNet-121 value of $0.5$) to yield the same number of channels ($256$) after the concatenation process. Similar to the CC-2P scheme, the PRM generated four patches (when $C_{p} = HC$) go through the C-stem and subsequent blocks one by one to yield their corresponding crowd counts that are summed to output the final crowd estimate for the input patch $P$ in this case. Empirically, it has been observed that the branching-out of the classification head after the DB2 achieves better results as compared to the same branching being deployed after other dense blocks as detailed in Sec. \ref{abl_subsection}.

\textbf{Proposed Approach and the Switch-CNN comparison.} Switch-CNN \cite{sam2017switching}, as detailed in Sec. \ref{intro123}, also classifies the input patch into the appropriate density level, followed by the crowd estimation using one of three specialized regressors. However, we approach this task in a totally different way by just using the straightforward plug-and-play PRM module with no learnable parameters and employing only one regressor or the base network. Whereas, the Switch-CNN uses complex coupling of the classifier with three specialized regressors. Consequently, the proposed frameworks (CC-Mod, CC-1P, CC-2P) have fewer model parameters ($9.85M,6.7M,5.05M$, respectively) as compared to the Switch-CNN ($15.1M$), and achieve better performance (see Sec. \ref{expi}).
\section{Evaluation and Training Details}

As per the standard crowd counting evaluation criteria, Mean Absolute Error (MAE) and Root Mean Square Error (RMSE) metrics have been used:
\vspace{-1mm}
\begin{equation}
\label{eq3}
MAE = \frac{1}{N} \sum_{k=1}^{N} |C_{k}-\hat{C_{k}}|, RMSE =\sqrt[]{ \frac{1}{N} \sum_{k=1}^{N} (C_{k}-\hat{C_{k}})^{2}}
\end{equation}

\noindent where $N$ indicates the total number of test images in the given benchmark dataset, and $C_{k}$, $\hat{C_{k}}$ represent the actual and the estimated counts respectively for test image $k$.

\textbf{Training Details.} In the end-to-end networks, the modular classifier and the crowd count regressor were trained separately using 90,000 patches each with mixed crowd numbers and original patch sizes of $112 \times 112$, $224 \times 224$, and $448 \times 448$. We used batch size of 16, stochastic gradient descent (SGD) as the optimizer and trains for 75 epochs with multi-step learning rate that starts at 0.001 and decreases to half each time after 25 and 50 epochs. Other parameters remain the same as for orignial DenseNet \cite{huang2017densely}. As per standard, $10\%$ training data has been used for model validation.

\section{Experimental Results}
\label{expi}
In this section, we report results obtained from extensive experiments on three diverse benchmark datasets: ShanghaiTech \cite{zhang2016single}, UCF-QNRF \cite{idrees2018composition}, and AHU-Crowd \cite{hu2016dense}. These datasets vary drastically from each other in terms of crowd diversity range, image resolution, and complex cluttered background patterns. First, we analyze standard quantitative experimental results and ablation study on these datasets, followed by the cross-dataset evaluation. In the end, we analyze some qualitative results.  

\setlength{\tabcolsep}{2.0pt}
\begin{table}[t]\small
	\begin{center}
	\begin{tabular}{|c|c|c|c|c|}
    \hline

 & \multicolumn{2}{c|}{ShanghaiTech} & \multicolumn{2}{c|}{UCF-QNRF}\\ \hline
Method & MAE  & RMSE & MAE  & RMSE\\ \hline
MCNN \cite{zhang2016single}  & 110.2  & 173.2 & 277   & 426   \\ \hline
Cascaded-MTL \cite{cascadedmtl} & 101.3  & 152.4 & 252   & 514   \\ \hline
Switch-CNN \cite{sam2017switching} & 90.4 & 135.0 & 228   & 445   \\ \hline
SaCNN \cite{zhang2018crowd} & 86.8 & 139.2 & -   & -   \\ \hline
IG-CNN \cite{babu2018divide} & 72.5 & 118.2  & -   & -  \\ \hline
ACSCP \cite{shen2018crowd}   & 75.7  & 102.7 & -   & - \\ \hline 
CSRNet \cite{li2018csrnet}   & 68.2  & 115.0 & -   & - \\ \hline 

CL\cite{idrees2018composition} & - & - & 132 & 191    \\ \hline

CFF \cite{shi2019counting} & 65.2 & 109.4 & \textbf{93.8}   & 146.5 \\ \hline
RRSP \cite{wan2019residual} & 63.1 & 96.2  & -   & -  \\ \hline 
CAN \cite{liu2019context} & \textbf{62.3} & 100.0 & 107   & 183   \\ \hline
TEDNet \cite{jiang2019crowd}  & 64.2  & 109.1 & 113   & 188  \\ \hline
L2SM \cite{xu2019learn} & 64.2 & 98.4 & 104.7   & 173.6 \\ \hline
Densenet121\cite{huang2017densely} & 93 & 139  & 167   & 229  \\ \hline \hline

CC-Mod (\textbf{ours}) & 73.8 & 113.2 & 107.8 & 171.2   \\ \hline
CC-2P (\textbf{ours}) & 67.8 & \textbf{86.2}  & 94.5 & \textbf{141.9}   \\ \hline
CC-1P (\textbf{ours})  & 69.1  & 109.5  & 97.3  & 153   \\ \hline  \hline

(CC-Mod/CC-2P/CC-1P) w/o PRM & 93.8 & 139.2 & 168 & 230     \\ \hline

	\end{tabular}
	\end{center}
	    \vspace{-1mm}
	\caption{\footnotesize ShanghaiTech \cite{zhang2016single} and UCF-QNRF \cite{idrees2018composition} datasets experiments and ablation study. Our PRM based approach (CC-2P) outperforms the state-of-the-art methods under the RMSE metric while giving competitive performance on MAE. Other PRM based proposed methods (CC-Mod and CC-1P) also give reasonable results. During the ablation study (last row), all proposed schemes give worse results after removing the PRM module, thus, indicating the quantitative importance of the proposed PRM. }
	\label{table:ST_results}
    \vspace{-5mm}
\end{table}

\subsection{ShanghaiTech Dataset Experiments}
\label{st_subsection}
The ShanghaiTech Part-A \cite{zhang2016single} dataset consists of diverse 482 images, with a predefined split of 300 training and  182 testing images. The proposed PRM based schemes are compared with the state-of-the-art methods as shown in Table \ref{table:ST_results}, where our approach (CC-2P) outperforms others under \textit{RMSE} evaluation metric with a significant improvement of $10.40\%$ ($96.2$ to $86.2$) and also give reasonable performance on MAE. The smallest RMSE value also indicates the lowest variance of our approach as compared with the other methods. Other proposed schemes (CC-Mod, CC-1P) also give comparable results in comparison.

To further evaluate the proposed methods, we removed the PRM module from each proposed scheme separately during the ablation study. After the PRM module removal, all three proposed schemes just become the same customized DenseNet based crowd count regressor (the base network), thus giving the same ablation performance as indicated by the last row of Table \ref{table:ST_results}. The ablation results show that the performance decreases dramatically (MAE: $27.71\%$, RMSE: $38.07\%$ error increase for CC-2P) without the PRM module, hence, quantifying the importance and effectiveness of the proposed PRM module.

\setlength{\tabcolsep}{2.0pt}
\begin{table}[t]\small
	\begin{center}
	\begin{tabular}{|c|c|c|c|}
    \hline
Method & MAE & RMSE\\ \hline
Haar Wavelet \cite{oren1997pedestrian} & 409.0   & -    \\ \hline
DPM \cite{felzenszwalb2008discriminatively} & 395.4  & -    \\ \hline
BOW–SVM \cite{csurka2004visual} & 218.8  & -    \\ \hline
Ridge Regression \cite{chen2012feature} & 207.4  & -    \\ \hline
Hu et al. \cite{hu2016dense} & 137  & -    \\ \hline

DSRM \cite{yao2017deep} &  81  &  129    \\ \hline

Densenet121\cite{huang2017densely} & 88.2  & 126.1    \\ \hline \hline

CC-Mod (\textbf{ours}) & 75.1 & 121.2    \\ \hline
CC-2P (\textbf{ours}) & \textbf{66.6} & \textbf{101.9}    \\ \hline
CC-1P (\textbf{ours})  & 70.3  & 107.2    \\ \hline  \hline

(CC-Mod / CC-2P / CC-1P) w/o PRM & 89.9 & 127    \\ \hline

	\end{tabular}
    \vspace{-2mm}
	\end{center}
	\caption{\footnotesize Our approach outperforms other models under all evaluation metrics on AHU-Crowd dataset. Ablation study (last row) also demonstrates the quantitative importance of the PRM module.}
	\label{table:AHU_results}
    \vspace{-1mm}
\end{table}

\setlength{\tabcolsep}{2.0pt}
\begin{table}[t]\small
	\begin{center}
	\begin{tabular}{|c|c|c|c|c|}
    \hline

 & \multicolumn{2}{c|}{ShanghaiTech} & \multicolumn{2}{c|}{UCF-QNRF}\\ \hline
Method & MAE  & RMSE & MAE  & RMSE\\ \hline

 \multicolumn{5}{|c|}{CC-Mod}\\ \hline
 Using VGG-16 & 79.3 & 125.6 & 128 & 181     \\ \hline
 Using VGG-19 & 78.9 & 124.1 & 122 & 179     \\ \hline
 Using ResNet-50 & 77.2 & 121.2 & 120 & 177     \\ \hline
 Using ResNet-101 & 77.0 & 120.5 & 121 & 176     \\ \hline
 Customized DenseNet-121 (ours) & \textbf{73.8} & \textbf{113.2} & \textbf{107.8} & \textbf{171.2} \\ \hline

 \multicolumn{5}{|c|}{CC-1P Branching-out}\\ \hline
 After DB1 & 78.3 & 123.2 & 128 & 181     \\ \hline
 After DB2 (ours) & \textbf{69.1} & \textbf{109.5} & \textbf{97.3} & \textbf{153}     \\ \hline
 After DB3 & 73.2 & 116.3 & 120 & 177     \\ \hline
 After DB4 & 79.1 & 124.2 & 121 & 176     \\ \hline
  
	\end{tabular}
	\end{center}
	\caption{\footnotesize Ablation Study on the CC-Mod architecture choice and the CC-1P Branching-out effect. The results justify our use of customized DenseNet-121 architecture as the 4-way classifier and the count regressor in the CC-Mod framework, and also our DB2 based branching-out selection of the Classification-Head in the CC-1P model.}
	\label{table:abl_results}
    \vspace{-1mm}
\end{table}

\setlength{\tabcolsep}{1.0pt}
\begin{table}\small 
	\begin{center}
	\begin{tabular}{|c|c|c|}
    \hline
Method & MAE  & RMSE\\ \hline
Cascaded-mtl \cite{cascadedmtl} & 308  & 478  \\ \hline
Switch-CNN \cite{sam2017switching} & 301  & 457   \\ \hline \hline

CC-Mod (\textbf{ours}) & 251 & 333    \\ \hline
CC-2P (\textbf{ours}) & \textbf{219} & \textbf{305}    \\ \hline
CC-1P (\textbf{ours})  & 227  & 318    \\ \hline

	\end{tabular}
	\end{center}
	    \vspace{-1mm}
	\caption{\footnotesize Cross-dataset evaluation. Models are trained using ShanghaiTech Part A images patches and tested on the UCF-QNRF dataset. Results show the generalization ability of the proposed method.}
	\label{table:transferLearning}
    \vspace{-1mm}
\end{table}

\subsection{UCF-QNRF Dataset Experiments}
UCF-QNRF \cite{idrees2018composition} dataset is the most diverse and challenging crowd counting benchmark dataset to date due to higher image resolutions, huge crowd diversity across different images and complex cluttered regions. It consists of 1,535 images with 1,251,642 annotations in total and a predefined training/testing split of 1,201/334 images, respectively. Also, the image resolution varies greatly from as low as $300 \times 377$ to as high as $6666 \times 9999$. As compared with the state-of-the-art models, our \textit{CC-2P} approach outperforms them under RMSE evaluation metric while performing reasonably closer in terms of RMSE, as shown in Table \ref{table:ST_results}. Our method shows a significant improvement as RMSE drops by $5.12\%$ ($146.5$ to $139$). The ablation study (last row of Table \ref{table:ST_results}), same as in Sec. \ref{st_subsection}, quantifies the importance of the PRM module (MAE: $43.75\%$, RMSE: $38.3\%$ error increase for CC-2P) after removing it from the proposed schemes.
\subsection{AHU-Crowd Dataset Experiments}
AHU-Crowd dataset \cite{hu2016dense} consists of 107 images and as per standard convention, we perform 5-fold cross-validation. As shown in Table \ref{table:AHU_results}, comparison based results show that our methods outperform other state-of-the-arts in terms of all evaluation criteria. Ablation study (last row of Table \ref{table:AHU_results}), same as in Sec. \ref{st_subsection}, demonstrates the importance and effectiveness of the PRM module.

\subsection{Ablation Study on CC-Mod Architecture Choice and CC-1P Branching-out Effect}
\label{abl_subsection}
In this experiment, we first explore different state-of-the-art architectures that can be used as the CC-mod classifier and the regressor. As shown in Table \ref{table:abl_results}, our customized DenseNet-121 choice performs the best in all cases. All other architectures are adapted for this ablation study as described in Sec. 5.6 of \cite{sam2017switching}. Next, we analyze the possible branching-out of the classification-head after each dense block (DB1, DB2, DB3, DB4) separately in CC-1P. Again, Table \ref{table:abl_results} justifies our DB2 based branching-out in CC-1P with least error.

\begin{figure*}
	\begin{minipage}[b][][b]{0.41\columnwidth}
		\begin{center}
			\centerline{\includegraphics[width=1\columnwidth]{figures/qual_results/IMG_30.jpg}}
			\centerline{\footnotesize{GT=307, DR=405}}
			\centerline{\footnotesize{Ours=306, DME\cite{zhang2018crowd}=451}}
		\end{center}
	\end{minipage}
		\begin{minipage}[b][][b]{0.41\columnwidth}
		\begin{center}
			\centerline{\includegraphics[width=1\columnwidth]{figures/qual_results/IMG_32.jpg}}
			\centerline{\footnotesize{GT=961, DR=1016}}
			\centerline{\footnotesize{Ours=949, DME\cite{zhang2018crowd}=1051}}
		\end{center}
	\end{minipage}
	\begin{minipage}[b][][b]{0.41\columnwidth}
		\begin{center}
			\centerline{\includegraphics[width=1\columnwidth]{figures/qual_results/IMG_40.jpg}}
			\centerline{\footnotesize{GT=236, DR=324}}
			\centerline{\footnotesize{Ours=246, DME\cite{zhang2018crowd}=299}}
		\end{center}
	\end{minipage}
	\begin{minipage}[b][][b]{0.41\columnwidth}
		\begin{center}
			\centerline{\includegraphics[width=1\columnwidth]{figures/qual_results/IMG_66.jpg}}
			\centerline{\footnotesize{GT=823, DR=948}}
			\centerline{\footnotesize{Ours=833, DME\cite{zhang2018crowd}=913}}
		\end{center}
	\end{minipage}			
		\begin{minipage}[b][][b]{0.41\columnwidth}
		\begin{center}
			\centerline{\includegraphics[width=1\columnwidth]{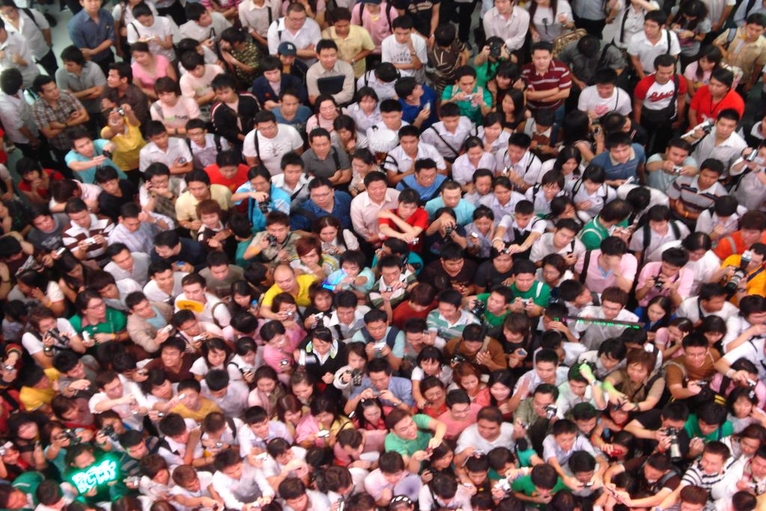}}
			\centerline{\footnotesize{GT=218, DR=275}}
			\centerline{\footnotesize{Ours=223, DME\cite{zhang2018crowd}=299}}
		\end{center}
	\end{minipage}	
	
		\begin{minipage}[c]{0.25\columnwidth}
		\begin{center}
			\centerline{\includegraphics[width=1\columnwidth]{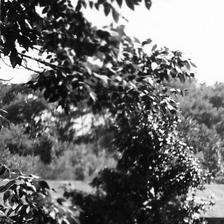}}
			\centerline{\footnotesize{NC}}
		\end{center}
	\end{minipage}
			\begin{minipage}[c]{0.25\columnwidth}
		\begin{center}
			\centerline{\includegraphics[width=1\columnwidth]{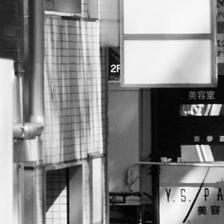}}
			\centerline{\footnotesize{NC}}
		\end{center}
	\end{minipage}
			\begin{minipage}[c]{0.25\columnwidth}
		\begin{center}
			\centerline{\includegraphics[width=1\columnwidth]{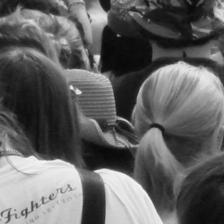}}
			\centerline{\footnotesize{LC}}
		\end{center}
	\end{minipage}
			\begin{minipage}[c]{0.25\columnwidth}
		\begin{center}
			\centerline{\includegraphics[width=1\columnwidth]{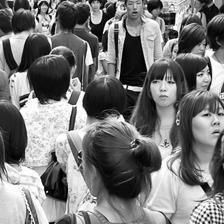}}
			\centerline{\footnotesize{LC}}
		\end{center}
	\end{minipage}	
		\begin{minipage}[c]{0.25\columnwidth}
		\begin{center}
			\centerline{\includegraphics[width=1\columnwidth]{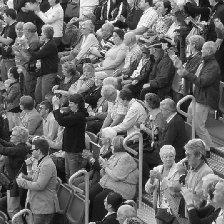}}
			\centerline{\footnotesize{MC}}
		\end{center}
	\end{minipage}
			\begin{minipage}[c]{0.25\columnwidth}
		\begin{center}
			\centerline{\includegraphics[width=1\columnwidth]{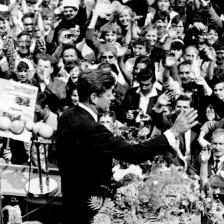}}
			\centerline{\footnotesize{MC}}
		\end{center}
	\end{minipage}
			\begin{minipage}[c]{0.25\columnwidth}
		\begin{center}
			\centerline{\includegraphics[width=1\columnwidth]{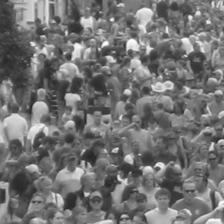}}
			\centerline{\footnotesize{HC}}
		\end{center}
	\end{minipage}
			\begin{minipage}[c]{0.25\columnwidth}
		\begin{center}
			\centerline{\includegraphics[width=1\columnwidth]{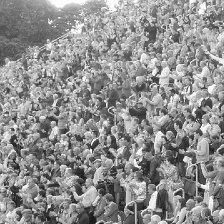}}
			\centerline{\footnotesize{HC}}
		\end{center}
	\end{minipage}	
						
    \vspace{-2mm}
	\caption{\footnotesize{ Qualitative results. First row shows the visual results from the actual test images. As compared with Direct Regression (DR) \cite{huang2017densely} and density-map estimation (DME) methods, our approach yields closer to the ground truth (GT) results. Second row shows our 4-way classification results, where it labels these complex patches correctly, thus, helping in routing the patches to the correct PRM rescaling option and also discards any no-crowd patch.
	}}
	\label{fig:qualityResults}
    \vspace{-4mm}
\end{figure*}

\subsection{Cross-dataset Evaluation}
We perform cross-dataset evaluation and compare the results with the state-of-the-art models. Each model is trained on the ShanghaiTech part-A training dataset and evaluated on UCF-QNRF dataset testing images. Results are shown in Table \ref{table:transferLearning}, where we compare our method with two state-of-the-art models. These results demonstrate that our approach is much more reliable with better generalization ability as it yields the best performance with a decrease in MAE (from $301$ to $219$) and RMSE (from $457$ to $305$).
\subsection{PRM Rescaling Operations Usage and Background Detection Analysis}
\begin{figure}[t]

	\begin{minipage}[b]{\columnwidth}
		\begin{center}
			\centerline{\includegraphics[width=0.985\columnwidth]{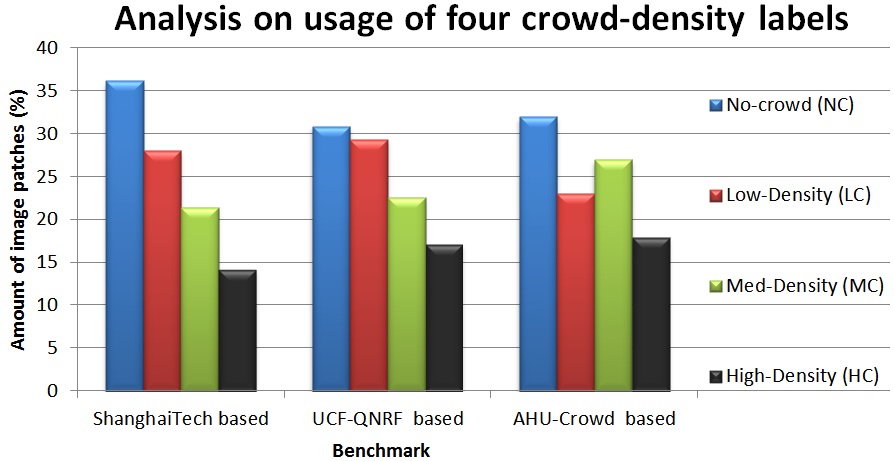}}
		\end{center}
	\end{minipage}		
    \vspace{-8mm}
	\caption{4-way Classification ($C_P$) results on each benchmark dataset reveal the frequency and importance of the PRM rescaling operations (as applied on LC and HC labeled patches). It also indicates that a large number of patches have been classified as no-crowd (NC) and thus discarded to avoid overestimation.}
    \vspace{-5mm}
    \label{fig:rescalersImportance}
\end{figure}

Here, we make a quantitative analysis of each PRM rescaling option usage as well as the amount of background (NC) patches being detected and discarded by the proposed scheme (CC-2P) during the test on three benchmark datasets. In all benchmark evaluations, either PRM rescaling option ($HC$ or $LC$) has been used for at least $14.25\%$ and as high as $29.4\%$ of the test images patches as shown in Fig. \ref{fig:rescalersImportance}. Thus, the PRM  have been utilized quite frequently and played an imperative role in enhancing the overall performance. Similarly, $36.2\%$, $30.9\%$ and $32\%$ (on average) of test image patches in ShanghaiTech, UCF-QNRF and AHU-Crowd datasets, respectively, have been detected as no-crowd (NC) and discarded after classification. These background patches could have created a great crowd overestimation otherwise as described in Sec. \ref{intro123}.

\subsection{Qualitative Results}
Some qualitative results have been shown in Fig. \ref{fig:qualityResults}, where the first row demonstrates the crowd count results on actual test images. As compared to Direct regression (DR) \cite{huang2017densely} and Density map estimation (DME) based methods, it is evident that our approach yields more accurate and reliable results. Sample crowd-density classification results, as shown in the second row of the same figure, demonstrate the effectiveness of the 4-way classification, which is crucial in routing the test patch to the correct PRM rescaling operation as well as in detecting any background image patch.

\section{Conclusion}
In this paper, we have presented an effective PRM module and three independent crowd counting frameworks. The proposed frameworks employ straightforward PRM rescaling operations instead of complex multi-column or multiple specialized crowd count regressors based architectures. The experimental results show that the proposed approach outperforms the state-of-the-art methods in terms of the RMSE metric and achieves competing performance in the MAE metric. The cross-dataset examination also indicates the great generalization ability of our method.

{\small
\bibliographystyle{ieee}
\bibliography{egbib}
}
\balance
\end{document}